\documentclass{ecai}

\PassOptionsToPackage{numbers, compress}{natbib}

\usepackage{graphicx}
\usepackage{latexsym}
\usepackage{amsmath}
\usepackage{amsthm}
\usepackage{booktabs}
\usepackage{natbib}
\usepackage{amsfonts}

\usepackage{multirow}
\usepackage{mathtools}

\begin{document}

\begin{frontmatter}

\title{Towards Flexible Time-to-event Modeling: Optimizing Neural Networks via Rank Regression}

\author[A,*]{\fnms{Hyunjun}~\snm{Lee}}
\author[B,*]{\fnms{Junhyun}~\snm{Lee}}
\author[C]{\fnms{Taehwa}~\snm{Choi}}
\author[B,\textdagger]{\fnms{Jaewoo}~\snm{Kang}\orcid{0000-0001-6798-9106}}
\author[D,\textdagger]{\fnms{Sangbum}~\snm{Choi}\orcid{0000-0001-6983-5821}}

\address[*]{Equal contribution}
\address[A]{Digital Tech. Center, SK Inc. C\&C}
\address[B]{Department of Computer Science and Engineering, Korea University}
\address[C]{Department of Biostatistics and Bioinformatics, Duke University}
\address[D]{Department of Statistics, Korea University}
\address[\textdagger]{Corresponding authors. Email: kangj@korea.ac.kr, choisang@korea.ac.kr}

\begin{abstract}
Time-to-event analysis, also known as survival analysis, aims to predict the time of occurrence of an event, given a set of features. 
One of the major challenges in this area is dealing with censored data, which can make learning algorithms more complex. 
Traditional methods such as Cox's proportional hazards model and the accelerated failure time (AFT) model have been popular in this field, but they often require assumptions such as proportional hazards and linearity.
In particular, the AFT models often require pre-specified parametric distributional assumptions.
To improve predictive performance and alleviate strict assumptions, there have been many deep learning approaches for hazard-based models in recent years.  
However, representation learning for AFT has not been widely explored in the neural network literature, despite its simplicity and interpretability in comparison to hazard-focused methods.
In this work, we introduce the Deep AFT Rank-regression model for Time-to-event prediction (\textit{DART}). This model uses an objective function based on Gehan's rank statistic, which is efficient and reliable for representation learning. 
On top of eliminating the requirement to establish a baseline event time distribution, \textit{DART} retains the advantages of directly predicting event time in standard AFT models.
The proposed method is a semiparametric approach to AFT modeling that does not impose any distributional assumptions on the survival time distribution. 
This also eliminates the need for additional hyperparameters or complex model architectures, unlike existing neural network-based AFT models. 
Through quantitative analysis on various benchmark datasets, we have shown that \textit{DART} has significant potential for modeling high-throughput censored time-to-event data.
\end{abstract}

\end{frontmatter}

%

\section{Introduction}
Time-to-event analysis, also known as survival or failure time analysis, is a widely used statistical method in fields such as biostatistics, medicine, and economics to estimate either risk scores or the  distribution of event time, given a set of features of subjects  \cite{vigano2000survival,cheng2016computer,dirick2017time,li2021difficulty}.
While assessing risk and quantifying survival probabilities have benefits, time-to-event analysis can be challenging due to the presence of censoring.
In real-world studies, subjects (e.g. patients in medical research) may be dropped out before the event of interest (e.g. death) occurs, which can prevent full follow-up of the data \cite{leung1997censoring}.
The presence of censoring in survival data can create a serious challenge in applying standard statistical learning strategies. 
In general, the censoring process is assumed to be non-informative in that it is irrelevant of the underlying failure process given features, but their relationship should be properly accounted for, otherwise leading to biased results.

Cox's proportional hazards (CoxPH) model is the most well-known method for time-to-event data analysis. 
It specifies the relationship between a conditional hazard and given features in a multiplicative form by combining the baseline hazard function with an exponentiated regression component, allowing for the estimation of relative risks.
However, this model requires so-called the proportional hazards assumption and time-invariant covariate-effects, which can be difficult to verify in many applications \cite{aalen1994effects}.
Statistical testing procedures, such as Schoenfeld's test, are typically conducted to examine the PH assumptions, as they are often vulnerable to violation of underlying assumptions \cite{aalen2001understanding,kleinbaum2010survival}.

The accelerated failure time (AFT) model, also known as the accelerated life model, relates the logarithm of the failure time to features in a linear fashion \cite{lee2003statistical}. 
This model has been used as an attractive alternative to the CoxPH model for analyzing censored failure time data due to its natural physical interpretation and connection with linear models.
Unlike the CoxPH models, the classical parametric AFT model assumes the underlying time-to-event distribution can be explained with a set of finite-dimensional parameters, such as Weibull or log-normal distribution. 
However, such assumption on the failure time variable can be restrictive and may not accurately reflect real-world data. 
This can decrease performance of the AFT model compared to Cox-based analysis, making it less attractive for practical use \cite{cox2008generalized,kleinbaum2010survival}.
Recently, researchers have been exploring a range of time-to-event models that leverage statistical theories and deep learning techniques to circumvent the necessity of assumptions such as linearity, single risk, discrete time, and fixed-time effect \cite{katzman2018deepsurv,lee2018deephit,ren2019deep,kvamme2019continuous,avati2020countdown,tarkhan2021survival,rahman2021deeppseudo}.

In particular, \textit{Cox-Time} \cite{kvamme2019time} and \textit{DATE} \cite{chapfuwa2018adversarial} alleviate some of the strict assumptions of the CoxPH and parametric AFT models by allowing non-proportional hazards and non-parametric event-time distribution, respectively.
\textit{Cox-Time} utilizes neural networks as a relative risk model to access interactions between time and features.
The authors also show that their loss function serves as a good approximation of the Cox partial log-likelihood.
\textit{DATE} is a conditional generative adversarial network that implicitly specifies a time-to-event distribution within the AFT model framework.
It does not require a pre-specified distribution in the form of a parameter, instead the generator can learn from the data using an adversarial loss function.
Incidentally, various deep learning-based approaches have been proposed to improve the performance by addressing issues such as temporal dynamics and model calibration \cite{lee2019dynamic,nagpal2021dcm,gao2021multi,kamran2021estimating,hu2021transformer}.
These approaches have highlighted the importance of utilizing well-designed objective functions that not only take into account statistical properties but also optimize neural networks.

In this paper, we introduce a Deep AFT Rank-regression for Time-to-event prediction model (\textit{DART}), a deep learning-based semiparametric AFT model, trained with an objective function originated from Gehan's rank statistic. 
We eliminate the need for specifying a baseline event time distribution while still preserving the advantages of AFT models that directly predict event times.
By constructing comparable rank pairs in the simple form of loss functions, the optimization of \textit{DART} is efficient compared to other deep learning-based event time models.
Our experiments show that \textit{DART} not only calibrates well, but also competes in its ability to predict the sequence of events compared to risk-based models.
Furthermore, we believe that this work can be widely applied in the community while giving prominence to the advantages of AFT models which are relatively unexplored compared to the numerous studies on hazard-based models.

\section{Related Works}

We first overview time-to-event modeling focusing on the loss functions of \textit{Cox-Time} and \textit{DATE} models to highlight the difference in concepts before introducing our method.
The primary interest of time-to-event analysis is to estimate survival quantities like survival function $S(t)=P(T \ge t)$ or hazard function $h(t)= \lim_{\delta \rightarrow 0}P(t \le T \le t + \delta|T \ge t)/\delta $, where $T \in \mathbb{R}^+$ denotes time-to-event random variable. 
In most cases, due to censored observations, those quantities cannot be directly estimated with standard statistical inference procedure.
In the presence of right censoring, \citeauthor{kaplan1958nonparametric} and \citeauthor{aalen1978nonparametric} provided consistent nonparametric survival function estimators, exploiting right-censoring time random variable $C \in \mathbb{R}^+$.  
Researchers then can get stable estimates for survival quantities with  data tuples $\{y_i, \delta_i,X_i\}_{i=1}^N$, where $y_i = \min (T_i, C_i)$ is the observed event time with censoring, $\delta_i = I(T_i \le C_i)$ is the censoring indicator, and a vector of features  $X_i \in \mathbb{R}^{P}$. 
Here, $N$ and $P$ denote the number of instances and  the number of features, respectively. 
While those nonparametric methods are useful, one can  improve predictive power by incorporating feature information in a way of regression modeling. 
Cox proportional-hazards (CoxPH) and accelerated-failure-time (AFT) frameworks are the most common approaches in modeling survival quantities utilizing both censoring and features.

\subsection{Hazard-Based Models}

A standard CoxPH regression model \cite{cox1972regression} formulates 
the conditional hazard function as:
\begin{equation}
h(t|X_i)=h_0(t)\exp(\beta^T X_i),~(i=1,\ldots,N),
\end{equation}
where $h_0(\cdot)$ is an unknown baseline hazard function which has to be estimated nonparametrically, and $\beta \in \mathbb{R}^{P}$ is the regression coefficient vector. 
It is one of the most celebrated models in statistics in that $\beta$ can be estimated at full statistical efficiency while achieving nonparametric flexibility on $h_0$ under the proportionality assumption. 
Note the model is semiparametric due to the unspecified underlying baseline hazard function $h_0$.
%
Letting $\mathcal{R}_i$ be the set of all individuals ``at risk'', meaning that are not censored and have not experienced the event before $T_i$, statistically efficient estimator for regression coefficients can be obtained minimizing the loss function with respect to $\beta$:
\begin{equation}\label{cox}
L_{\text{CoxPH}}(\beta) = \sum_{i}\delta_i\log\left( \sum_{j\in \mathcal{R}_i}  \exp\left[\beta^T X_j - \beta^T X_i\right] \right)
\end{equation}
which is equivalent to the negative partial log-likelihood function of CoxPH model.

Based on this loss function, \citeauthor{kvamme2019time} proposed a deep-learning algorithm for the hazard-based predictive model, namely \textit{Cox-Time}, replacing $\beta^TX_j$ and $\beta^TX_i$ with $g(y_j, X_j; \theta)$ and $g(y_i, X_i; \theta)$, respectively. 
Here, $g(\cdot)$ denotes the neural networks parameterized by $\theta$, and $\mathcal{R}_i$ would be replaced by $\tilde{\mathcal{R}}_i$, representing the sampled subset of $\mathcal{R}_i$. 
With a simple modification of the standard loss function in Eq. (\ref{cox}), \textit{Cox-Time} can alleviate the proportionality for relative risk,
showing empirically remarkable performance against other hazard-based models in both event ordering and survival calibration.
\subsection{Accelerated-Failure-Time Models}

The conventional AFT model relates the log-transformed survival time to a set of features in a linear form:
\begin{equation} \label{aft_model}
\log T_i  = \beta^T X_i + \epsilon_i,~(i=1,\ldots,N),
\end{equation}
where $\epsilon_i$ is an independent and identically distributed error term with a common distribution function $F_0(\cdot)$ that is often assumed to be  Weibull, exponential, log-normal, etc.
As implied in Eq. (\ref{aft_model}), AFT model takes a form of linear modeling and provides an intuitive and physical interpretation on event time without detouring via the vague concept of hazard function, making it a powerful alternative to hazard-based analysis. 
However, imposing a parametric distributional assumption for $\epsilon_i$ is a critical drawback of the model, for which model in Eq. (\ref{aft_model}) could be a subclass of the hazard-based models. 

To alleviate linearity and parametric distributional assumptions, several works brought the concept of generative process and approximated the error distribution via neural networks like generative adversarial networks (GANs) \cite{miscouridou2018deep,chapfuwa2018adversarial}. 
Especially, \citeauthor{chapfuwa2018adversarial} proposed a deep adversarial time-to-event (\textit{DATE}) model, which specifies the loss function as:
\begin{align}
\label{DATE}
    L_{\text{DATE}}(\theta, \phi) &=  \mathbb{E}_{(X, y) \sim F_{nc}}[D_\phi(X, y)]   \\
    +& \mathbb{E}_{X\sim F_{nc}, \xi\sim F_{\xi}}[1 - D_\phi(X, G_\theta(X, \xi; \delta = 1))] \nonumber \\ 
    +& \lambda_2 \mathbb{E}_{(X, y)\sim F_{c}, \xi \sim F_{\xi}}[\max(0, y - G_\theta(X, \xi; \delta = 0))] \nonumber \\
    +& \lambda_3 \mathbb{E}_{(X, y) \sim F_{nc},\xi \sim F_{\xi}}[\| t - G_\theta(X, \xi; \delta = 1) \|_1] \nonumber
\end{align}
where $\theta$ and $\phi$ denote the parameter set 
associated with a generator $G_\theta$ and a discriminator $D_\phi$, respectively,
$(\lambda_2, \lambda_3)$ are hyperparameters to tune censoring trade-off, 
$F_{nc}(X, y)$ and  $F_{c}(X, y)$ are empirical joint distributions for non-censored cases and censored cases, respectively, and $F_{\xi}$ is the simple distribution, such as uniform distribution.
The generator $G_\theta$ implicitly defines event time distribution.
Despite \textit{DATE} achieves prominent survival calibration via the sample-generating process, the objective function is quite complicated and the GAN framework is inherently prone to mode collapse, i.e., the generator learns only a few modes of the true distribution while missing other modes \cite{srivastava2017veegan}.
Also, when optimizing neural networks with multiple loss functions, it is difficult to balance and there might be conflicts (i.e. trade-off) with each term \cite{Dosovitskiy2020You}.
Therefore, their loss function might be difficult to be optimized as intended and requires a burdening training time, and consequently not be suitable for large-scale time-to-event analysis.

In the statistical literature, there have been many attempts to directly estimate regression coefficients in the semiparametric AFT model, where  the error distribution $F_0$ is left unknown, rather than imposing specific parametric distribution or  exploiting generative models. 
In this work, we bridge non-linear representation learning and an objective function for estimation of semiparametric AFT model, which is originated from Gehan's rank statistic.
By extensive quantitative analysis, we have shown the beauty of simplicity and compatibility of rank-based estimation, along with outstanding experimental performance.

\begin{figure*}[t]
\centerline{\includegraphics[width=0.756\textwidth]{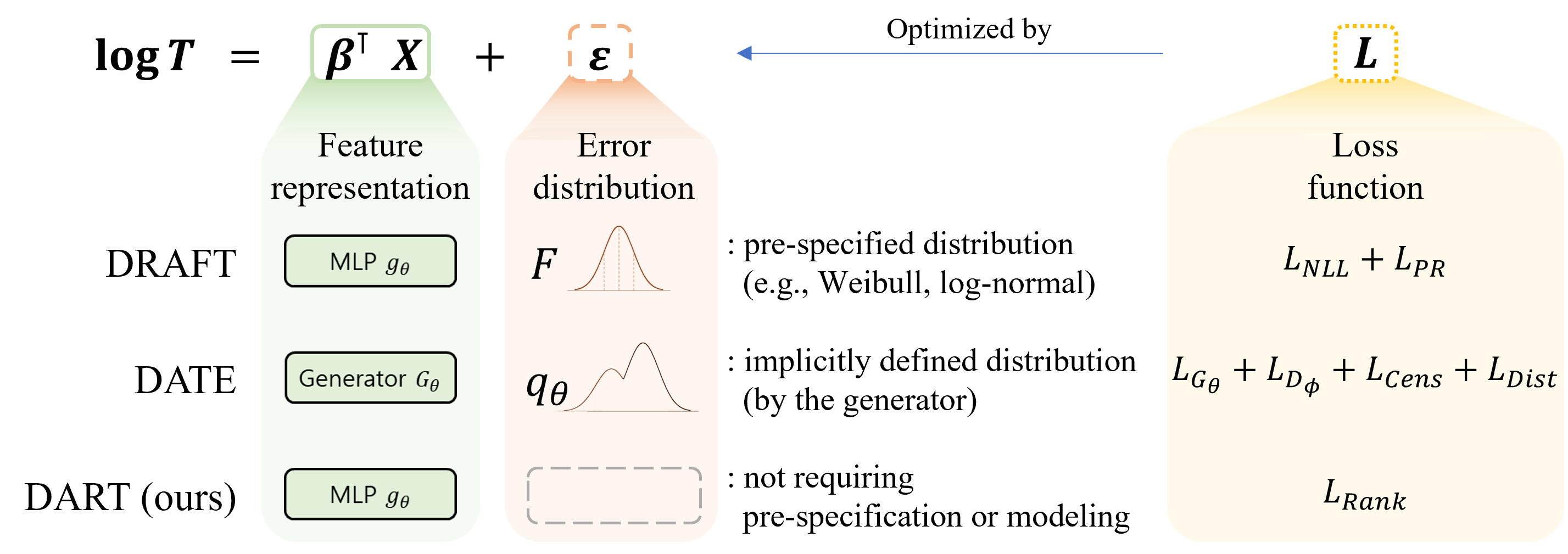}}
\caption{Illustration of conceptual differences between deep learning-based AFT models in terms of their respective contributions and required assumptions with a format of the standard AFT. 
To alleviate the parametric distribution assumption, which \textit{DRAFT} has, \textit{DATE} exploits the GAN framework and learns the implicit underlying distribution $q_\theta$ through the generator parameterized by $\theta$. 
For \textit{DRAFT}, $L_{NLL}$ and $L_{PR}$ denote negative log-likelihood and partial ranking likelihood, respectively.
\textit{DATE} basically requires four loss functions: $L_{G_{\theta}},L_{D_{\phi}}$ for the generator and the discriminator, $L_{Cens}$ for adjusting censoring distribution, and $L_{Dist}$ for the distortion penalty. 
Compared to the others, \textit{DART} does not require pre-specification or modeling for error distribution and it is trained with a simple loss function supported by statistical theory.}
\label{figure_1}
\end{figure*}

\section{Method}


In this section, we introduce the concept of \textit{DART}, followed by predictive analysis for survival quantities.
The conceptual differences with the other neural network-based AFT models are illustrated in Figure \ref{figure_1}.
The semiparametric AFT is distinct from a parametric version in that the error distribution function $F_0$ is left completely unknown like the baseline hazard function in the CoxPH.
By further exploiting neural networks, we propose \textit{DART} model that can be formulated as a  generalization of model in Eq. (\ref{aft_model}):
\begin{equation} \label{DART}
    \log T_i = g(X_i; \theta) + \epsilon_i,~(i=1,\ldots,N),
\end{equation}
where $g(X_i; \theta)$ denotes arbitrary neural networks with input feature vector $X_i$ and a parameter set $\theta$, outputting single scalar value as predicted log-scaled time-to-event variable. 
With this simple and straightforward modeling, \textit{DART} entails several attractive characteristics over existing AFT-based models. 
First, the semiparametric nature of \textit{DART} enables flexible estimation of error distribution, allowing improved survival prediction via neural network algorithms for $F_0$.
Second, the restrictive log-linearity assumption of AFT model can be further alleviated by exploiting deep neural networks.
Specifically, while standard AFT model relates time-to-event variable to feature variable in linear manner, deep learning is able to approximate any underlying functional relationship, lessening linearity restriction. 
Although \textit{DART} still requires log-transformed time as a target variable, its deep neural network redeems the point with powerful representative performance supported by universal approximation theorems, enabling automated non-linear feature transformation \cite{leshno1993multilayer,schafer2006recurrent,zhou2020universality}.


\subsection{Parameter Estimation via Rank-based Loss Function}

In statistical literature,  many different estimating techniques have been proposed for fitting semiparametric AFT model \cite{tsiatis1990estimating,jin2003rank,jin2006least,zeng2007efficient}. 
Among them, we shall adopt the $l_1$-type rank-based loss function by taking into account the censoring information, which is efficient and suitable for stably fitting neural networks.
Letting a residual term $e_i\equiv e_i(\theta)=\log y_i - g(X_i;\theta)$, the objective loss function for \textit{DART} can be formulated as:
\begin{equation}\label{objfun}
    \displaystyle
    L_{\text{Rank}}(\theta)=\cfrac{1}{N}\sum_{i=1}^N\sum_{j=1}^N \delta_i (e_i - e_j) I\{e_i \ge e_j\},
\end{equation}
where $I(\cdot)$ is the indicator function that has value 1 when the condition is satisfied, otherwise 0.
The estimator $\hat{\theta}$ can be obtained by minimizing the loss function with respect to model parameter set $\theta$. 
Optimization of model parameters can be conveniently conducted via batched stochastic gradient descent (SGD).
Notice that the loss function in Eq. (\ref{objfun}) involves model parameter $\theta$ only, without concerning estimation of the functional parameter $F_0$, enabling flexible time-to-event regression modeling. 

Strength of the loss function is theoretical consistency of optimization without requiring any additional settings. 
Let the neural network be expressed: $g(X_i; \phi, \beta) = \beta^T W_i$, where $W_i \in \mathbb{R}^{K}$ is transformed feature vector through hidden layers with parameter set $\phi$, and $\beta \in \mathbb{R}^{K}$ is a parameter set of linear output layer. 
Then, it is easy to see that the following estimating function is the negative gradient of the loss function with respect to $\beta$: 
\begin{align}
\label{esteq}
U_{\text{Rank}}(\beta) &= \frac1N \sum_{i=1}^N\sum_{j=1}^N \delta_i(W_i-W_j) I(\tilde{e}_i \le \tilde{e}_j) \overset{\text{set}}{=} 0 \nonumber \\
\tilde{e}_i &= \log y_i - \beta^T W_i.
\end{align}
Eq. (\ref{esteq}) is often called the form of Gehan's rank statistic \cite{jin2003rank}, testing whether $\beta$ is equal to true regression coefficients for linear model $\log T_i = \beta^T W_i + \epsilon_i$, and the solution to the estimating equation $\hat\beta$ is equivalent to the minimizer of Eq. (\ref{objfun}) with respect to $\beta$. 
This procedure entails nice asymptotic results such as $\sqrt n$-consistency and asymptotic normality of $\hat\beta$ under the counting processes logic, assuring convergence of $\hat\beta$ towards true parameter $\beta$ as the number of instances gets larger \cite{tsiatis1990estimating,jin2003rank}.
Although these asymptotic results might not be directly generalized to the non-linear predictor function, we expect that hidden layers would be able to assess effective representations $W_i$ with non-linear feature transformation, as evidenced by extensive quantitative studies. 
Note that, to keep theoretical alignment, it is encouraged to set the last layer as a linear transformation with an output dimension of 1 to mimic the standard linear model following non-linear representation.
In addition, a robust estimation against outlying instances can be attained, depending rank of residual terms along with their difference.

\subsection{Prediction of Survival Quantities}

Predicted output $g(X_i; \hat{\theta})$ from trained \textit{DART} model represents estimated expectation of $\log T_i$ conditional on $X_i$, i.e. mean log-transformed survival time with given feature information of $i$th instance. 
However, estimating survival quantities (e.g. conditional hazard function) cannot be directly done for AFT-based models. 
Instead, we utilize the Nelson-Aalen estimator \cite{aalen1978nonparametric}, verified to be consistent under the rank-based semiparametric AFT model \cite{park2003estimating}.
Define $N(t;\theta) = \sum_{i=1}^N N_i(t)$ and $Y(t)=\sum_{i=1}^N Y_i(t)$, 
where $N_i(t) = I( e_i \le t, \delta_i = 1 )$ and $Y_i(t)=I(e_i > t)$ are the counting and the at-risk processes, respectively.
Then the Nelson-Aalen estimator of $H_0(t)$ is defined by
\begin{equation}
\hat H_0(t) = \int_0^t \dfrac{ I\{Y(u)>0\} }{Y(u)} dN(u).
\end{equation}
The resulting conditional hazard function given $X_i$ is defined by
\begin{equation}
\hat{h}(t|X_i) = \hat{h}_0[t\exp\{-g(X_i; \hat{\theta})\}]\exp\{-g(X_i;\hat{\theta})\},
\end{equation}
where $\hat{h}_0(\cdot) = d\hat H_0(\cdot) $ is pre-trained baseline hazard function using Nelson-Aalen estimator. 
Consequently, conditional survival function can be estimated by relationship $\hat{S}(t|X_i) = \exp\{-\int_0^t \hat{h}(t|X_i)dt\}$, providing comparable predictions to other time-to-event regression models. 
In practice, training set is used to get pre-trained Nelson-Aalen estimator.



\begin{table*}[h]
\caption{Summary of survival datasets.}
\vskip -0.15in
\label{table_1}
\begin{center}
\begin{sc}
\begin{tabular}{lrrr}
\toprule
Dataset & Size & \# features & \% Censored \\
\midrule
WSDM KKBox     &2,646,746 & 15  & 0.28\\
SUPPORT      & 8,873 & 14 & 0.32 \\
FLCHAIN      & 6,524 & 8  & 0.70 \\
GBSG & 2,232 & 7  & 0.43 \\
\bottomrule
\end{tabular}
\end{sc}
\end{center}
\vskip -0.1in
\end{table*}

\section{Evaluation Criteria}
In this section, we evaluate models with two metrics for quantitative comparison: concordance index (CI) and integrated Brier score (IBS).

\textbf{Concordance Index.}
Concordance of time-to-event regression model represents the proposition: if a target variable of instance $i$ is greater than that of instance $j$, then the predicted outputs of $i$ should be greater than that of $j$. 
By letting target variable $y$ and predicted outcome $\hat{y}$, concordance probability of survival model can be expressed as $P(\hat{y}_i > \hat{y}_j|y_i > y_j)$, and concordance index measures the probability with trained model for all possible pairs of datasets \cite{harrell1982evaluating}.
With non-proportional-hazards survival regression models like \textit{Cox-Time} or \citet{lee2018deephit}, however, \citet{harrell1982evaluating} cannot be used to measure discriminative performance properly. 
For fair comparison of survival regression models, time-dependent concordance index \cite{antolini2005time}, or $C^{\text{td}}$ was used for those baseline models proposed by \citet{kvamme2019time} to account for tied events. 
$C^{\text{td}}\in[0,1]$ can be regarded as AUROC curve for time-to-event regression model, denoting better discriminative performance for a value close to 1. 
Note that standard concordance index yields identical results with $C^{\text{td}}$ for AFT-based models.

\textbf{Integrated Brier Score.}
\citet{graf1999assessment} introduced generalized version of Brier score \cite{brier1950verification} for survival regression model along with inverse probability censoring weight (IPCW), which can be described as:
\begin{equation}
\begin{split}
    \text{BS}(t) = \cfrac{1}{N}\sum_{i=1}^N 
    \cfrac{\hat{S}(t|X_i)^2 I(y_i \le t, \delta_i = 1)}{\hat{G}(y_i)} \\ +\cfrac{1}{N}\sum_{i=1}^N \cfrac{(1 - \hat{S}(t|X_i))^2 I(y_i > t)}{\hat{G}(t)}
\end{split}
\end{equation}
where $\hat{G}(t)=\hat{P}(C>t)$ is a Kaplan-Meier estimator for censoring survival function to assign IPCW. 
$\text{BS}(t)$ measures both how well calibrated and discriminative is predicted conditional survival function: if a given time point $t$ is greater than $y_i$, then $\hat{S}(t|X_i)$ should be close to 0.
Integrated Brier score (IBS) accumulates BS for a certain time grid $[t_1, t_2]$:
\begin{equation}
    \text{IBS} = \cfrac{1}{t_2 - t_1}\int_{t_1}^{t_2}BS(s)ds.
\end{equation}
If $\hat{S}(t|X_i)=0.5$ for all instances, then $\text{IBS}$ becomes 0.25, thus well-fitted model yields lower IBS.
For experiments, time grids can practically be set to minimum and maximum of $y_i$ of the test set, equally split into 100 time intervals.


\begin{table}
\caption{Hyperparameter search space for the WSDM KKBox dataset.}
\vskip -0.15in
\label{tab:hyper_kkbox}
\begin{center}
\begin{tabular}{lccc}
\toprule
Hyperparameter &Values \\
\midrule
\# Layers &\{4,6,8\} \\
\# Nodes per layer & \{128, 256, 512\}   \\
Dropout & \{0.0, 0.1, 0.5\} \\
\bottomrule
\end{tabular}
\end{center}
\vskip -0.1in
\end{table}

\begin{table}
\caption{Hyperparameter search space for GBSG, FLCHAIN, and SUPPORT datasets.}
\vskip -0.15in
\label{tab:hyper_others}
\begin{small}
\begin{center}
\begin{tabular}{lccc}
\toprule
Hyperparameter &Values \\
\midrule
\# Layers &\{1, 2, 4\} \\
\# Nodes per layer & \{64, 128, 256, 512\}   \\
Dropout & \{0.1, 0.2, 0.3, 0.4, 0.5, 0.6, 0.7\} \\
Weight decay & \{0.4, 0.2, 0.1, 0.05, 0.02, 0.01, 0.001\}\\
Batch size & \{64, 128, 256, 512, 1024\}\\
$\lambda$ (CoxTime and CoxCC) & \{0.1, 0.01, 0.001, 0.0\} \\
\bottomrule
\end{tabular}
\end{center}
\end{small}
\vskip -0.1in
\end{table}

\section{Experiments}

In this section, we describe our experiment design and results to validate performance of \textit{DART} compared to other time-to-event regression models. 
We conduct experiments using four real-world survival datasets and baseline models provided by \citeauthor{kvamme2019time} and \citeauthor{chapfuwa2018adversarial} with two evaluation metrics mentioned in previous section. 

\begin{table*}
\caption{Mean and standard deviation of $C^{\text{td}}$. 
HAZ and AFT denote hazard-based and AFT-based methods, repectively.}
\vskip -0.15in
\label{table_2}
\begin{center}
\begin{tabular}{clcccc}
\toprule
\multicolumn{2}{c}{MODEL}  & WSDM KKBox  & SUPPORT & FLCHAIN& GBSG  \\
\midrule
\multirow{3}{*}{\rotatebox{90}{HAZ}}&\textit{DeepSurv} &0.841 (0.000)  & 0.619	(0.008)  & 0.797	(0.013)& 0.685 (0.013)\\
&\textit{Cox-CC}   & 0.836 (0.046)& 0.618	(0.009) & 0.797(0.013) & 0.684 (0.012) \\
&\textit{Cox-Time} & \textbf{0.853} (0.049)& \textbf{0.637}	(0.009)   &\textbf{0.800}	(0.012)& \textbf{0.687} (0.012)\\
\midrule
\multirow{3}{*}{\rotatebox{90}{AFT}} &\textit{DRAFT} & 0.861 (0.005)  & 0.599	(0.018)& 0.725	(0.057)& 0.611 (0.016)  \\
&\textit{DATE} & 0.852 (0.001)  & 0.608	(0.008)& 0.784	(0.009) & 0.598 (0.034)  \\
&\textit{DART (ours) }    & \textbf{0.867} (0.001)&  \textbf{0.624}	(0.009) & \textbf{0.797}	(0.014) &\textbf{0.687} (0.014) \\
\bottomrule
\end{tabular}
\end{center}
\end{table*}

\begin{table*}[t]
\caption{Mean and standard deviation of Integrated Brier Score (IBS).}
\vskip -0.15in
\label{table_3}
\begin{center}
\begin{tabular}{clcccc}
\toprule
\multicolumn{2}{c}{MODEL}  & WSDM KKBox & SUPPORT & FLCHAIN &  GBSG \\
\midrule
\multirow{3}{*}{\rotatebox{90}{HAZ}}&\textit{DeepSurv} &0.111 (0.000) &  \textbf{0.190}	(0.004) & \textbf{0.101}	(0.006) & \textbf{0.174}	(0.004)\\
&\textit{Cox-CC}   & 0.115 (0.012) &  0.191	(0.003) & 0.122	(0.028) &0.177	(0.004) \\
&\textit{Cox-Time} & \textbf{0.107} (0.009) & 0.194	(0.006) & 0.114	(0.016) & \textbf{0.174}	(0.005) \\
\midrule
\multirow{3}{*}{\rotatebox{90}{AFT}} &\textit{DRAFT} & 0.147 (0.002) & 0.314	(0.043) &0.144	(0.022) & 0.310	(0.010) \\
&\textit{DATE} & 0.131 (0.002) & 0.227	(0.004) & 0.124	(0.012) & 0.204	(0.004) \\
&\textit{DART (ours) } &\textbf{0.108} (0.001) &\textbf{0.176}	(0.005)  &  \textbf{0.068}	(0.007)&  \textbf{0.150}	(0.023) \\
\bottomrule
\end{tabular}
\end{center}
\vskip -0.1in
\end{table*}


\subsection{Datasets}
We use three benchmark survival datasets and a single large-scale dataset provided by \citeauthor{kvamme2019time}.
The descriptive statistics are provided in Table \ref{table_1}.
Specifically, three benchmark survival datasets include: 
the Study to Understand Prognoses Preferences Outcomes and Risks of Treatment (SUPPORT), the Assay of Serum Free Light Chain (FLCHAIN), and the Rotterdam Tumor Bank and German Breast Cancer Study Group (GBSG). 
Of particular interest is the GBSG dataset, which includes an indicator variable for hormonal therapy, allowing us to evaluate the effectiveness of a treatment recommendation system built using survival regression models. 
In addition, we use the large-scale WSDM KKBox dataset containing more than two millions of instances for customer churn prediction, which was prepared for the 11th ACM International Conference on Web Search and Data Mining. 
With a large-scale dataset, we can clearly verify the consistency of predictive performance of time-to-event models.

\subsection{Baseline Models}
We select six neural network-based time-to-event regression models as our experimental baselines: \textit{DRAFT} and \textit{DATE} \cite{chapfuwa2018adversarial} as AFT-based models for direct comparison with our model, and \textit{DeepSurv} \cite{katzman2018deepsurv},  \textit{Cox-CC} and \textit{Cox-Time} \cite{kvamme2019time} as hazard-based models.

For AFT-based models, \textit{DRAFT} utilizes neural networks to fit log-normal parametric AFT model in non-linear manner. 
However, it should be noted that if the true error variable does not follow a log-normal distribution, this model may be misspecified.
In contrast, \textit{DATE} exploits generative-adversarial networks (GANs) to learn conditional time-to-event distribution and censoring distribution using observed dataset. 
The generator's distribution of \textit{DATE} can be trained from data to implicitly encode the error distribution.
One major benefit of this approach is that it eliminates the need to pre-specify the parameters of the distribution.
Despite these advantages, the method is challenging to apply to real-world datasets due to the complexity of the training procedure and objective function.

In case of hazard-based models, \textit{DeepSurv} fits Cox regression model whose output is estimated from neural networks. 
The model outperforms the standard CoxPH model in performance, not clearly exceeding other neural network-based models.
Furthermore, the proportional hazards assumption still remains unsolved with \textit{DeepSurv}.
\textit{Cox-CC} is another neural network-based Cox regression model, using case-control sampling for efficient estimation. 
While both \textit{DeepSurv} and \textit{Cox-CC} are bounded to proportionality of baseline hazards, \textit{Cox-Time} relieves this restriction using event-time variable to estimate conditional hazard function.

In this study, we focus on neural network-based models and exclude other machine learning-based models from the comparison to avoid redundant analysis that was conducted in previous studies. 
Some neural network-based models are also excluded as we aim to alleviate fundamental assumptions such as proportionality and parametric distribution. 
Note that comparing hazard-based models and AFT-based models has been rarely studied due to their difference in concept and purpose: modeling hazard function and modeling time-to-event variable. 
While models can be evaluated using common metrics, it is important to conduct a thorough analysis when analyzing numerical experiments, particularly when comparing hazard-based models and AFT-based models. 
These two types of models have different underlying concepts and purposes, making it crucial to take into consideration their unique characteristics during the analysis.

\begin{table*}[t]
\caption{Comparison of the training time (seconds) per epoch over the KKBox dataset.}
\vskip -0.15in
\label{time}
\begin{center}
\begin{tabular}{ccccccc}
\toprule
 &  \textit{DeepSurv} & \textit{Cox-CC} & \textit{Cox-Time} & \textit{DRAFT} & \textit{DATE} & \textit{DART (ours)} \\ \midrule
Time &27.81 & 44.86 & 42.60 & 759.04 & 2024.19 & 29.93 \\
\bottomrule
\end{tabular}
\end{center}
\end{table*}

\subsection{Model Specification and Optimization Procedure}
For a fair comparison, we apply neural network architecture used in \citeauthor{kvamme2019time}: MLP with dropout and batch-normalization. 
Every dense blocks are set to have the equal number of nodes (i.e. the dimension of hidden representations), no output bias is utilized for output layer, and ReLU function is chosen for non-linear activation for all layers.
Preprocessing procedure has also been set based on \citeauthor{kvamme2019time} including standardization of numerical features, entity embeddings \cite{guo2016entity} for multi-categorical features. 
The dimension of entity embeddings is set to half size of the number of categories.
In addition, due to the fact that parameters of AFT-based models tend to be influenced by scale and location of the target variable, $y$ has been standardized and its mean and variance are separately stored to rescaled outputs.

\textbf{\textit{DeepSurv, Cox-CC, Cox-Time, DART}.} 
The PyCox\footnote{https://github.com/havakv/pycox} python package provides the training codes for these models.
For WSDM KKBox dataset, we repeat experiments 30 times with best configurations provided by \citeauthor{kvamme2019time}.
Because train/valid/test split of KKBox dataset is fixed, we don't perform a redundant search procedure.
For the other datasets (SUPPORT, FLCHAIN, and GBSG), we perform 5-fold cross-validation as performed at \citeauthor{kvamme2019time} because the size of datasets is relatively small.
At each fold, the best configuration is selected among 300 combinations of randomly selected hyperparameters which are summarized in Table \ref{tab:hyper_others}.
We use AdamWR \cite{loshchilov2017decoupled} starting with one epoch of an initial cycle and doubling the cycle length after each cycle.
The batch size is set to 1024 and the learning rates are found by \citeauthor{smith2017cyclical} as performed at \citeauthor{kvamme2019time}.

\textbf{\textit{DRAFT, DATE}.} 
The implementation of \textit{DATE}\footnote{https://github.com/paidamoyo/adversarial\_time\_to\_event} by the authors includes the code of \textit{DRAFT} as well.
We utilize their official codes for all datasets.
The batch size for KKBox dataset is set to 8192 because the experiments are not feasible with the batch size 1024 due to their training time.
The best configurations of \textit{DRAFT} and \textit{DATE} also are founded by grid search with same hyperparameter search space with others.
We repeat experiments 30 times with the best configuration as mentioned above.
For the other datasets, as same with other models, we perform 5-fold cross-validation and choose the best configuration among 300 random hyperparameter sets at each fold.
The hyperparameter search space for WSDM KKBox dataset is summarized in Table \ref{tab:hyper_kkbox}.
Our implementation for \textit{DART} is publicly available at: \url{https://github.com/teboozas/dart_ecai23}.

\subsection{Performance Evaluation}
To measure discriminative performance of outputs, we exploit standard C-index \cite{harrell1982evaluating} for AFT-based models while letting hazard-based models to utilize $C^{\text{td}}$ since equivalent evaluation is possible for AFT-based models including \textit{DART} since it outputs a single scalar value to evaluate ranks.
In terms of survival calibration, we implement our own function to obtain IBS based on its definition, due to the fact that evaluation methods of the conditional survival function and IPCW provided by \citeauthor{kvamme2019time} are not compatible with AFT-based models. 
Specifically, we first fit Kaplan-Meier estimator upon standardized training set, and subsequently evaluate conditional survival estimates and IPCW utilizing estimated residuals, following the definition of baseline hazard function of AFT framework rather than to use time-to-event variable directly.
For numerical integration, we follow settings of time grid from \citeauthor{kvamme2019time}, and standardize the grid with mean and standard deviation stored with standardization procedure of training set. 
By doing so, IBS can be compared upon identical timepoints for both hazard-based models and AFT-based models.

\subsection{Summary of Results}

Experiment results are provided in Table \ref{table_2} and \ref{table_3}. 
In summary, \textit{DART} is competitive in both discriminative and calibration performance, especially for large-scale survival datasets. 
Specifically, \textit{DART} yields consistent results for WSDM KKBox dataset compared to other baselines, maintaining competitive performance in terms of $C^{\text{td}}$ and IBS. 
We point out that \textit{DART} is the most powerful and AFT-based time-to-event model that can be a prominent alternative when hazard-based models might be not working.

\section{Analysis}
We provide analysis on experimental results, pointing out strengths of \textit{DART} model in terms of performance metrics.

\textbf{Characteristic of \textit{DART} for large-scale dataset.} As provided in Table \ref{table_2} and \ref{table_3}, \textit{DART} generally yields prominent survival calibration performance with small variance in terms of IBS.
Especially for large-scale dataset (KKBox), \textit{DART} shows state-of-the-art performance with the smallest variance in evaluated metrics.
This result comes from the characteristic of rank-based estimation strategy.
Specifically, on the basis of asymptotic property of Eq. (\ref{esteq}), estimated model parameters get stable and close to true parameter set, when the size of dataset gets larger.
Thus, once the trained model attains effective representation ($W_i$ in Eq. (\ref{esteq})) from hidden layers via stochastic optimization methods, \textit{DART} is able to provide stable outputs with strong predictive power, without sophisticated manipulation upon time-to-event distribution.


\textbf{Comparison with AFT-based models.} In case of \textit{DRAFT}, model does not generally perform well for both $C^{\text{td}}$ and IBS for most datasets. 
This is attributed to the fact that \textit{DRAFT} is a simple extension of the parametric AFT model with log-normality assumption. 
Thus, this approach is quite sensitive to true underlying distribution of dataset. 
On the other hand, \textit{DATE} yields clearly improved performance against \textit{DRAFT} especially for survival calibration in terms of IBS. 
Unlike \textit{DRAFT}, \textit{DATE} utilizes GAN to learn conditional error distribution without parametric assumption, allowing the model to yield more precise survival calibration. 
However, time-to-event distribution is trained with divided loss functions by optimizing two tuning hyperparameters in Eq. (\ref{DATE}).
This approach can be significantly affected by well-tuned hyperparameters and heavy computation is required to this end, resulting insufficient performance.
Meanwhile, as illustrated in Figure \ref{figure_1}, \textit{DART} has advantages of simplicity in theoretical and practical points compared to the other AFT-based models.

\textbf{Comparison with hazard-based models.} As previously reported by \citeauthor{kvamme2019time}, \textit{Cox-Time} shows competitive performance against other hazard-based models, directly utilizing event-time variable to model conditional hazard function.
However, we found out that \textit{Cox-Time} requires precise tuning of additional hyperparameters ($\lambda$ and Log-durations) largely affecting predictive performance.

In contrast, \textit{DART} shows smaller variance in evaluation metrics as the size of data increases, ensuring stable output for large-scale dataset with asymptotic property which is crucial for practical application.
In addition, while \textit{Cox-Time} showed better performance against \textit{DART} with respect to C-index in some cases, \textit{DART} outperformed in \textit{IBS}; gaining equivalent mean IBS scores against \textit{Cox-Time} with smaller variance indicates our method dominant others in comprehensive survival metrics.


\textbf{Comparison of the required time for optimizing each model.}
To verify the compatibility for large-scale data, we measure the training time of each model.
We strictly bound the scope of the target process for a fair comparison, as from data input to parameter update excluding other extra steps.
The specifications of all models are set equally: the number of nodes 256, the number of layers 6, and the batch size 1024.
With the consumed time of 1000 iterations, we calculate the training time for a single epoch.
We exclude the first iteration that is an outlier in general.
All experiments were run on a single NVIDIA Titan XP GPU.
Table \ref{time} shows that the simplicity of \textit{DART} leads to practical efficiency, while \textit{DATE} is computationally expensive due to the generator-discriminator architecture.

\textbf{Notes on practical impact of performance gain}
We acknowledge that practically interpreting IBS might be less intuitive and challenging. Thus we provide a simplified example below. Consider that a random-guess model, which estimates all conditional survival functions at 0.5, would result in an IBS score of 0.250. In this context, an improvement from 0.174 to 0.150, which are close to IBS of \textit{Cox-Time} and \textit{DART} for GBSG case respectively, of is indeed substantial.

{\footnotesize
\begin{center}
\begin{tabular}{|c c|c|}
  \hline
  Perfect estimation & Random guess & IBS \\
  \hline
  0 & 100 & 0.250 \\
  30 & 70 & 0.175 \\
  40 & 60 & 0.150 \\
  \hline
\end{tabular}
\end{center}
}

While the analogy below might not be entirely suitable, one can infer the practical improvement in survival estimation precision.
It is worth to mention that the decreased IBS from 0.175 to 0.150 is still significant improvement in model accuracy, even though this comparison is somewhat rough.
It represents an increase in \textit{Perfect} estimations (i.e. $\hat{S}(t|X_i)=0\ \forall i \ \text{s.t. } t \ge y_i, \delta_i = 1$) from 30 to 40 occurrences (+33.3\%) out of 100 estimations. 
Considering the fact that $\hat{S}(t|X_i)$ ranges from 0 to 1, it is hard to achieve 0.025 points improvements in real settings.
Consequently, while the differences in the metrics might appear moderate, we would like to emphasize that they are practically significant.
Regarding the C-index, our model showed a 1.64\% improvement compared to Cox-Time on the KKBox dataset, with scores rising from 0.853 to 0.867.
Considering that a perfect score is 1.00, this implies that our model shows noticeable performance and exhibits high consistency
In summary, \textit{DART} would be the attractive alternative to existing time-to-event regression frameworks by ensuring remarkable performance and fast computation

\section{Conclusion}
In this work, we propose flexible time-to-event regression model, namely  \textit{DART}, utilizing the semiparametric AFT rank-regression method, coupled with  deep neural networks, to alleviate strict assumptions and to attain practical usefulness in terms of high and stable predictive power. 
Extensive experiments have shown that our approach is prominent in discrimination and correction performance even on large-scale survival datasets.
Although we do not yet address more complex censoring data such as competing risks and interval censoring, our approach can provide a stable baseline to handle these tasks in the near future with simple modifications of the loss function.

\section*{Acknowledgements}
The research conducted by Sangbum Choi was supported by grants from the National Research Foundation of Korea (2022M3J6A1063595, 2022R1A2C1008514) and the Korea University research grant (K2305261).
Jaewoo Kang and Junhyun Lee's research was supported by grants from the National Research Foundation of Korea (NRF-2023R1A2C3004176), the Korea Health Industry Development Institute (HR20C0021(3)), and the Electronics and Telecommunications Research Institute (RS-2023-00220195).

\bibliography{ecai.bib}
\end{document}